\def\year{2024}\relax
\documentclass[letterpaper]{article} 
\usepackage{aaai24}  
\usepackage{times}  
\usepackage{helvet}  
\usepackage{courier}  
\usepackage[hyphens]{url}  
\usepackage{graphicx} 
\usepackage{natbib}  
\usepackage{caption} 
\usepackage[ruled,vlined]{algorithm2e}
\SetKwProg{Function}{function}{}{}

\nocopyright
\usepackage{amsfonts}
\usepackage{amsmath}
\usepackage{stmaryrd}

\newcommand{\llbr}[1]{{\llbracket{#1}\rrbracket}}

\newcommand{\popX}{\mathcal{X}}
\newcommand{\popY}{\mathcal{Y}}
\newcommand{\adjFact}{\mathit{adj}}
\newcommand{\overX}{\widehat{X}}
\newcommand{\overY}{\widehat{Y}}
\newcommand{\overW}{\widehat{W}}
\newcommand{\overBeta}{\widehat{\beta}}
\newcommand{\ovx}{\widehat{x}}
\newcommand{\ovw}{\widehat{w}}

\title{
Using Stratified Sampling to Improve LIME Image Explanations
}
\author{
    Muhammad Rashid\textsuperscript{\rm 1},
    Elvio G. Amparore\textsuperscript{\rm 1},
    Enrico Ferrari\textsuperscript{\rm 2},
    Damiano Verda\textsuperscript{\rm 2}
}
\affiliations{
    \textsuperscript{\rm 1}University of Torino,~ Computer Science Department,~
    C.so Svizzera 185,~ 10149 Torino,~ Italy\\
    \textsuperscript{\rm 2}Rulex Innovation Labs,~ 
    Via Felice Romani 9,~ 16122 Genova,~ Italy\\
    \{muhammad.rashid, elviogilberto.amparore\}@unito.it,
    \{enrico.ferrari, damiano.verda\}@rulex.ai
}

\begin{document}

\maketitle

\begin{abstract}
We investigate the use of a stratified sampling approach for LIME Image, a popular model-agnostic explainable AI method for computer vision tasks, in order to reduce the artifacts generated by typical Monte Carlo sampling.
Such artifacts are due to the undersampling of the dependent variable in the synthetic neighborhood around the image being explained, which may result in inadequate explanations due to the impossibility of fitting a linear regressor on the sampled data.
We then highlight a connection with the Shapley theory, where similar arguments about undersampling and sample relevance were suggested in the past.
We derive all the formulas and adjustment factors required for an unbiased stratified sampling estimator. 
Experiments show the efficacy of the proposed approach.
\end{abstract}

\section{Introduction}

The efficacy of explainable AI techniques for computer vision tasks has seen several important advancements in the recent years. Several methods to interpret model predictions have emerged, as surveyed for instance by \cite{liang2021explaining} or \cite{guidotti2018survey}. 
In this paper we inspect the sampling strategy of one of these methods known as \emph{LIME Image}\,\cite{ribeiro2016should}, which is a \emph{model-agnostic} method (i.e. it is not tied to a particular type of black box model being explained) that produces \emph{feature attributions} as explanations.
As the name suggests, LIME Image is a method specialized for image classification tasks, and the ``feature attribution'' are importance scores assigned to regions of an input image 
measuring how much each region contributes to the model classification.

Feature attributions are the regression coefficients that solve a weighted least squares problem on a sampled population denoted as \emph{synthetic neighborhood}. 
Since the sampling process is inherently stochastic, the synthetic neighborhood may be inadequate for LIME Image to fit the regressor, resulting in slow convergence\,\cite{visani2022statistical} or instability\,\cite{sevillano2022revel}.
Sometimes, the explanation produced by LIME Image fails to identify any relevant region, resulting in regression coefficients with very small and almost uniform values (i.e. with low \emph{variation}, as we shall see).
We review the LIME Image process, focusing on the limitation of using a Monte Carlo sampling for the synthetic neighborhood generation. 


\medskip
\paragraph{Paper Contributions.}
In this paper we:
\begin{itemize}
    \item investigate the distribution of the dependent variable 
    in the sampled synthetic neighborhood of LIME Image, identifying in the undersampling a cause that results in inadequate explanations;

    \item delve into the causes of the synthetic neighborhood inadequacy, recognizing a link with the Shapley theory;
    
    \item reformulate the synthetic neighborhood generation 
    using an \emph{unbiased stratified sampling} strategy;

    \item provide empirical proofs of the advantage of using 
    stratified sampling for LIME Image on a popular dataset.

\end{itemize}

\begin{figure*}
    \centering
    \includegraphics[width=1.0\textwidth]{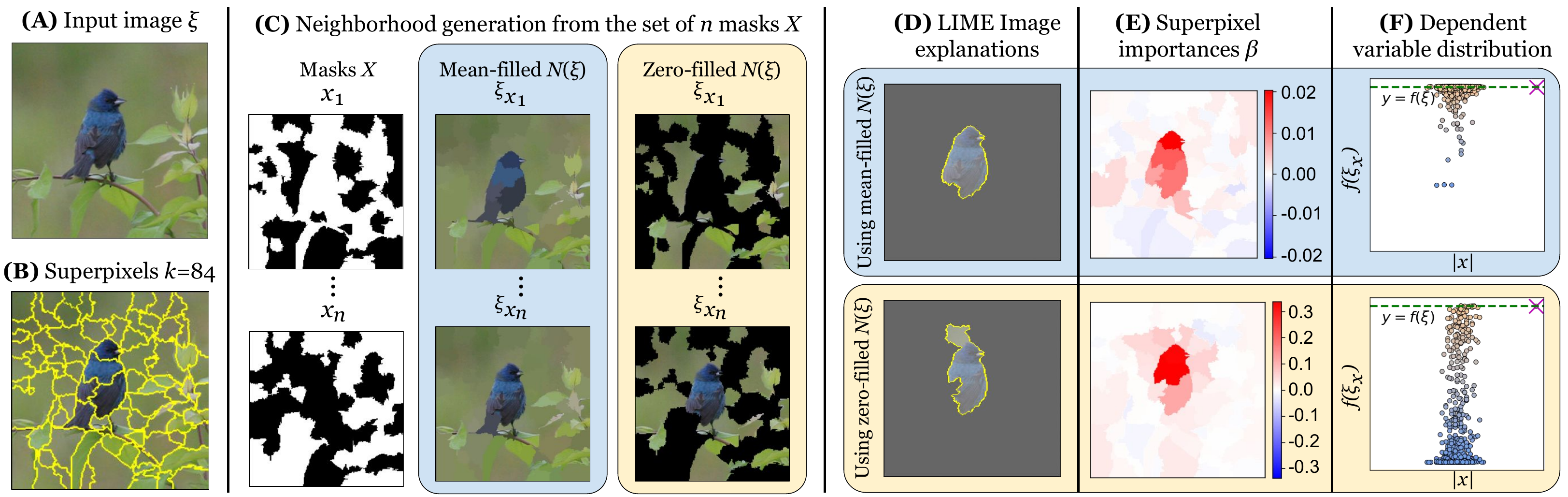}
    \caption{LIME Image workflow.}
    \label{fig:limeWorkflow}
\end{figure*}

\section{Previous Work}

A relevant theoretical study of LIME Image is\,\cite{garreau2021does}, which we partially summarize in the \emph{Preliminaries} section for the sake of self-containment, that also focuses on connections with \emph{integrated gradients}. 
Discretization of the synthetic neighborhood for tabular data has been studied in\,\cite{garreau2020explaining}, and for text data by\,\cite{mardaoui2021analysis}.
However, the setting for image data is significantly different, since the sample space is Boolean and not continuous. 
Sampling strategies received more attention in the context of the Shapley theory\,\cite{lundberg2017unified}, as in\,\cite{mitchell2022sampling}. 
We recast some of the intuitions of these previous works in the context of LIME, particularly from the \emph{multilinear extensions} \cite{owen1972multilinear}.

Several alternative sampling strategies for LIME have been studied. 
A clique-based sampling was considered in\,\cite{shi2020extension}. Moreover, sampling variance has been considered in several articles like \cite{zhang2019should} or in \cite{shankaranarayana2019alime}, where standard deviations of Ridge coefficients are compared.
A complementary study about region flipping analysis in LIME explanations is \cite{ng2022towards}, which could also be used to improve the approach proposed in this paper.
To the best of our knowledge, we are not aware of a consistent framework that adds unbiased stratified sampling to LIME.


\section{Preliminaries}

We briefly review how LIME works for image inputs, in order to explain our changes and their effects.
Fig.\,\ref{fig:limeWorkflow} depicts the LIME Image workflow steps, and will be used throughout this section to provide examples.
Consider the domain of RGB images of size $h \times w$, denoted as $\mathcal{I} \in [0-255]^{h \times w \times 3}$.
Let $f : \mathcal{I} \rightarrow \mathbb{R}$ be a black-box regression model function that provides a prediction score given an input image\footnote{We consider only the case of a binary class prediction, as the multi-class prediction is usually treated as several one-vs-rest binary class prediction problems.}, and let $\xi \in \mathcal{I}$ be the sample image being explained.
The main purpose of LIME is to generate a \emph{linear} model $g$ that locally approximate the explained black-box model $f$ in the neighborhood of an input sample $\xi$.

LIME explanations are not build directly on the image $\mathcal{I}$, but on a smaller domain denoted as the \emph{interpretable representation}. 
This domain is obtained by divided the input image into $k$ \emph{superpixels}\,(also called \emph{segments}, \emph{regions} or \emph{patches}) using an algorithm like \emph{quick shift} \cite{vedaldi2008quickshift}.
A superpixel is a contiguous region of pixels of $\xi$ that share some kind of similarity, and such that the $k$ superpixels form a partition of the pixels of $\xi$.
Fig.\,\ref{fig:limeWorkflow}A shows an example of an image taken from\,\cite{imagenet-object-localization-challenge}.
\,\ref{fig:limeWorkflow}B shows its segmentation obtained from the \emph{quick shift} algorithm\footnote{Using: $\mathit{kernel\_size}=4, \mathit{max\_dist}=7, \mathit{ratio}=0.2$.}, 
resulting in $k=84$ superpixels.
The model being used for the classification is ResNet50\,\cite{resnet50he2016deep}, pretrained for the ImageNet task. The image in Fig.\,\ref{fig:limeWorkflow}A is correctly classified as \emph{indigo\_bunting} with probability $99.49\%$.

The approach of LIME Image is based on the concept of \emph{superpixel masking}.
Let $x \in \{0,1\}^k$ be a binary vector (mask) representing the presence (value 1) or the absence (value 0) of each of the $k$ superpixels. 
Giving a mask $x$, a \emph{perturbed input image} $\xi_x$ is obtained by preserving the pixels of each superpixel $i$ having $x[i]=1$, and replacing every other pixel whose superpixel $i$ has $x[i]=0$.
Replacement can be done in several ways.
By default pixels of a masked superpixel $i$ are replaced by the mean color of that superpixel
(\emph{mean-filled}). 
Alternatively, they can be replaced with a fixed color value, like black (\emph{zero-filled}).
We use notation $x' = x[i\gets v]$ to denote a new mask $x'$ obtained from a mask $x$ by replacing the value for superpixel $i$ with $v$. Moreover, let $|x|$ be the number of preserved superpixels, i.e. those having $x[i]=1$.


In LIME Image, the individual values of a mask vector $x$ are sampled using an unbiased Monte Carlo strategy, i.e.
\begin{equation}\label{eq:maskB}
    x[i] \sim B(0.5),
    \qquad
    1 \leq i \leq k
\end{equation}
where $B(p)$ is a Bernoulli-distributed random variable having probability $p{=}0.5$.
A \emph{set of masks} $X$ with $n$ samples is made by randomly sampling $n$ instances of \eqref{eq:maskB} for the same input image $\xi$ having $k$ superpixels.
A \emph{synthetic neighborhood} $N(\xi) = \{ \xi_x ~|~ x \in X \}$ with $n$ samples is made by perturbing the input image $\xi$ using $n$ randomly sampled masks.
A depiction of the set of $n$ masks is shown in Fig.\,\ref{fig:limeWorkflow}C: randomly sampled masks $x_i$ are used to generate perturbed input images $\xi_{x_i}$, using two replacement strategies.

All the perturbed samples $N(\xi)$ can be classified by the black-box model $f$, resulting in the \emph{dependent variables}
\begin{equation}
    Y = \bigl\{ f(\xi_x) ~\big|~ \xi_x \in N(\xi) \bigr\}
\end{equation}

A \emph{distance function} is adopted, in order to weight the perturbed samples differently.
The intuition followed by LIME is that samples closer to $\xi$ should weight more.
\\
Given a mask $x$, the weight $w_x$ is
\begin{equation}\label{eq:weight}
    w_x = \exp \left( \frac{-D(x)^2}{\sigma^2} \right)
\end{equation}
where $D$ is the cosine similarity score between $x$ and $\vec{\mathbf{1}}$ (the vector of ones, i.e. the mask where everything is preserved), while $\sigma=0.25$ (by default) is the \emph{kernel width}. 
See \cite{garreau2020explaining} for an analysis on the role of Eq.\,\eqref{eq:weight} and of $\sigma$.
In this paper we will use the default value, as the focus is in the sampling methodology.
Let $W = \{ w_x ~|~ x \in X \}$.

Having the matrices of the set of masks $X \in \{0,1\}^{n \times k}$, the weights $W \in \mathbb{R}^{n \times 1}$ and the dependent variables $Y \in \mathbb{R}^{n \times 1}$ for all the observed samples in the synthetic neighborhood $N(\xi)$, then $Y$ can be written as the response variable of the \emph{linear regression model}.
LIME adopts a \emph{simple linear homoscedastic model}\,\cite{dumouchel1983using} for its regression coefficients, which is
\begin{equation}\label{eq:linReg}
    Y = X \cdot \beta + \epsilon
\end{equation}
where the vector $\beta$ is the weighted least squares estimator of the regression coefficients of $Y$ on $X$ weighted by $W$.

To simplify our analysis, we will consider no regularization factors (default for LIME Image is ridge regression with $L^2$ regularization), similarly to \cite{garreau2020explaining}.
This simplification does not affect significantly the main observations of this paper, which is focused on the sampling strategy.
The coefficients $\beta$ results from
\begin{equation}\label{eq:beta}
    \beta = (X^\textsf{T} W X)^{-1} X^\textsf{T} W Y
\end{equation}
which solves Eq.\,\eqref{eq:linReg}.
A linear function $g(x)$  with coefficients $\beta$ is a linear regressor that locally approximates the initial black-box model $f$.

\begin{figure}[t]
    \centering
    \includegraphics[width=0.9\columnwidth]{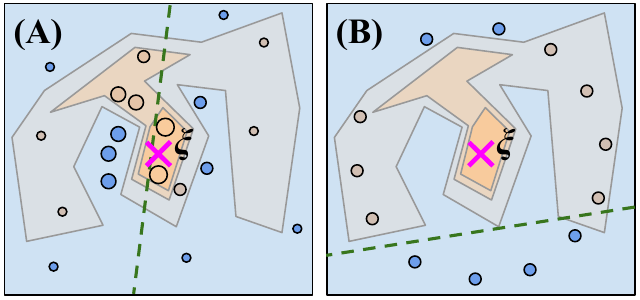}
    \caption{How LIME is supposed to work (A), and how it actually works (B) using Monte Carlo sampling for a large enough $k$.}
    \label{fig:limeNeighborhood}
\end{figure}

\paragraph{Interpretation of LIME.}

The $k$ coefficients of $\beta$ can be interpreted as \emph{feature importances} (or \emph{feature attributions}) of each of the $k$ superpixels of the input image $\xi$.
In that sense, the $k$ superpixels form the set of \emph{interpretable features} of the input image, over which the explanation is built.

There are two levels of interpretation of $\beta$. 
By default LIME Image suggests to select only the superpixels with the highest value (Fig.\,\ref{fig:limeWorkflow}D), resulting in an sub-region in the image (the {\small\textsf{get\_image\_and\_mask}} method). The number of selected superpixels is decided by the user: LIME does not provide an heuristic for this task. 
Alternatively, the coefficients can be visualized as an \emph{heatmap}, identifying the contribution of each superpixel to the classification (Fig.\,\ref{fig:limeWorkflow}E). 
The color intensity represents the value, with white representing the zero. 
Coefficients with higher absolute values means that the corresponding superpixel is more important in the classification outcome $f(\xi)$.
The scale of the coefficients can vary (in Fig.\,\ref{fig:limeWorkflow}E the same scale is used for both heatmaps) and it is known to not be particularly relevant \cite[pag.\,6]{garreau2020explaining} (only the ratios among the coefficients is).

Finally, it is relevant to inspect the distribution of the $Y$ values in the neighborhood (i.e. the values of $f(\xi_x)$) with respect to the count $|x|$ of masked superpixels (Fig.\,\ref{fig:limeWorkflow}F). 
This plot shows if the $Y$ values are sampled across the entire distribution (top to bottom), or if there are clear unbalances. 
In Fig.\,\ref{fig:limeWorkflow}F, the distribution for the zero-filled case has a good balance, since there are values obtained from the black box model $f$ covering the whole spectrum of values, while in the plot for the mean-filled case the balance is problematic, having most $Y$ values concentrated in the top.
As we shall see in the next section, imbalances in this distribution results in poor explanations being generated by LIME Image.

\section{Limitations of LIME Image Sampling}

While there has been a number of successful applications of LIME\,\cite{bodria2023benchmarking}, the explanation process largely depends on several factors. 
One such factors is the sampling process, which is stochastic and inherently uncertain. 
The use of a Monte Carlo strategy in Eq.~\eqref{eq:maskB} to sample the interpretable feature space when it is made by more than a few dozen of superpixels has important consequences.

\paragraph{Under-Representation of the Neighborhood.}
The intuition behind LIME is depicted in Fig.\,\ref{fig:limeNeighborhood}A, which is inspired by the one found in \cite[~Fig.\,3]{ribeiro2016should}. The explained sample $\xi$ (represented as a cross) is surrounded by its synthetic neighborhood $N(\xi)$ (represented as dots), whose classifications are obtained by the black box model $f$ and weighted by their proximity to $\xi$ (size of dots). 
A linear regressor (the green dashed line) is fit on these points weighted by their distance to $\xi$, and in principle it should be locally faithful to $f(\xi)$.
LIME Image however works like that only when the number of superpixels is very small. 
Since masks are obtained from Eq.\,\eqref{eq:maskB} having a fixed Bernoulli coefficient of $0.5$, the probability of selecting a mask $x$ having a given number of preserved superpixels $|x|$ follows the binomial distribution $\mathcal{B}(k, |x|)$ with probability mass function $\binom{k}{|x|}p^{|x|} (1-p)^{k-|x|}$.

\begin{figure}[t]
    \centering
    \includegraphics[width=\columnwidth]{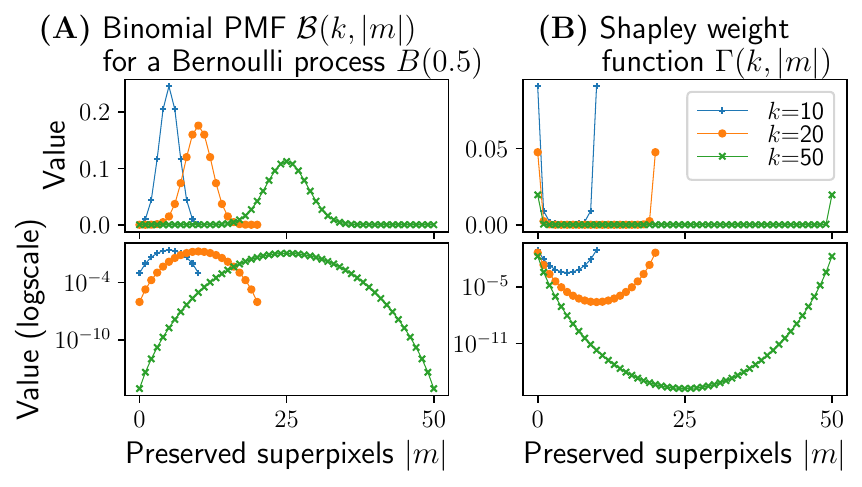}
    \caption{Binomial (A) and Shapley weight (B) distributions for $k = 10, 20$ and $50$.}
    \label{fig:binomShapley}
\end{figure}


Fig.\,\ref{fig:binomShapley}A shows the probability mass function for a few $k$ values, being $k$ the number of superpixels. 
This PMF is of course not uniform, and the probability of randomly sample points at the extremes drops rapidly. There is no indication of how many superpixels LIME Image can manage, but both the default parameters and practical experience\,\cite{vermeire2022explainable} shows that an image needs to be split into tens or even a few hundreds of superpixels, in order to have enough patches to correctly identify object borders. 
In that case, samples will distribute around $\xi$ forming a sort of hypersphere, as illustrated in Fig.\,\ref{fig:limeNeighborhood}B, where almost no sample is really close to $\xi$, since the probability of the binomial distribution concentrates around samples having ${\sim}50\%$ of the superpixels masked. 
In that way, the local behaviour (i.e. samples with $|x|$ close to $k$) is under-represented in the neighborhood.

\paragraph{Dependent Variables Distribution.}

As seen in Fig.\,\ref{fig:binomShapley}A, by increasing the superpixels $k$ the probability of getting samples from the tails of the distribution is practically reduced to $0$.
This effect depends on both the model and the input image: if selecting randomly about $50\%$ of the superpixels still allows the model to produce a ``reasonable'' distribution of the dependent variable $Y$, a linear regressor can be fit and an explanation can be produced. If however the $Y$ distribution is flattened, no reasonable explanation can be produced, as the linear regressor will be fit on almost uniform values.

\begin{figure}[t]
    \centering
    \includegraphics[width=1.0\columnwidth]{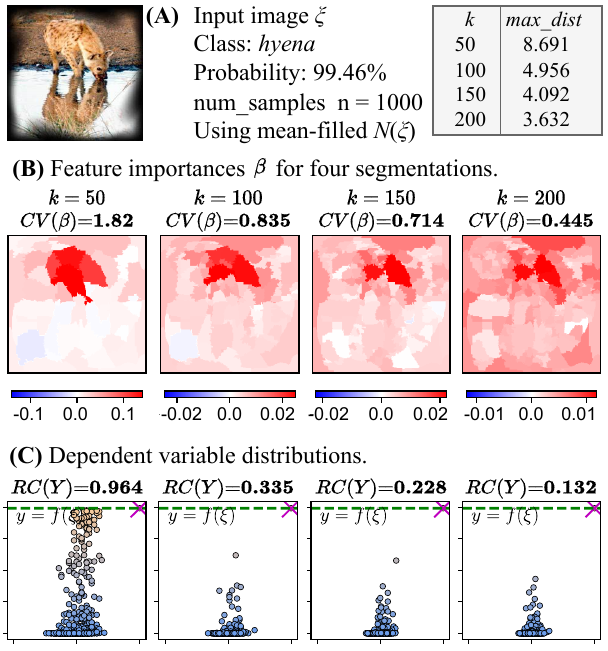}
    \caption{Dependent variable undersampling (low $RC(Y)$) results in confused explanations (low $CV(\beta)$).}
    \label{fig:hyenaMC}
\end{figure}

Fig.\,\ref{fig:hyenaMC} shows an example of this behaviour.
The input image \textbf{(A)} is correctly classified by the model as \emph{hyena} with high probability.
Feature importance vectors $\beta$ and the distribution of the dependent variables $Y$ (versus the number of masked superpixels $|x|$) are shown in \textbf{(B)} and \textbf{(C)}, respectively, for four different segmentations ($k=50, 100, 150$ and $200$ superpixels, respectively).
All values (heatmaps, $CV(\beta)$, $RC(Y)$) are averages of $10$ computations, to reduce randomicity in the reported results.
With $k=50$ segments (left), the $Y$ distribution has enough variability to obtain a vector $\beta$ that highlights which segments are more important.
Increasing the number of superpixels reduces such variability in the $Y$ distribution, resulting in explanations that are more and more ``confused''.
On these distributions it is of course harder to fit a linear regressor that is truthful to the explanation. 
Intuitively, it is like Fig.\,\ref{fig:limeNeighborhood}B where the hypersphere is almost entirely far away from $\xi$. In that case, the explanation produced by LIME Image will be progressively more meaningless.

In these problematic cases the values of the $\beta$ vector also drops to very small numbers (scale is reported below each heatmap in \textbf{(B)}), and variability across the feature importances decreases.
To quantitatively measure such form of ``confusion'', we employ the standard \emph{coefficient of variation}, defined as
\begin{equation}\label{eq:CV}
    CV(\beta) = \frac{\sigma_{\beta}}{\mu_{\beta}}
\end{equation}
where $\sigma_\beta$ and $\mu_\beta$ are the standard deviation and the mean of $\beta$, respectively.
Ideally, a good $CV(\beta)$ should not be close to zero (which would mean that all superpixels have almost the same value, and no clear sub-region in the image is identified).
The $CV(\beta)$ values for the example in Fig.\,\ref{fig:hyenaMC} are reported in the \textbf{(B)} row.

We also want to quantify the (approximate) \emph{range coverage} of the $Y$ values in the synthetic neighborhood. 
Theoretically this range is $[0, f(\xi)]$, but of course it can have under- or over-shoots due to the nature of the classification model.
To do so, we measure the proportion of that range that is contained in the $1\%-99\%$ interquantile range (IQR) of the $Y$ distribution, using
\begin{equation}\label{eq:RCY}
    RC(Y) = \frac{IQR_{1-99}(Y)}{f(\xi)}
\end{equation}
Low values of $RC(Y)$ indicate that the sampled $Y$ distribution is squashed into a small range of values, not covering the full $[0, f(\xi)]$ spectrum (like in Fig.\,\ref{fig:hyenaMC}C/right). 
Ideally $RC(Y)$ should be far from zero to have a good coverage of the probability range $[0, f(\xi)]$ by $Y$.

\paragraph{Sample Relevance.}
In the recent years, the Shapley theory\,\cite{lundberg2017unified} has received a lot of attention in the context of model-agnostic explainability, due to its flexibility and its axiomatic formulation\,\cite{rozemberczki2022shapley}.
While LIME does not have a corresponding axiomatic definition, we can still learn some insights from how Shapley values are defined over a weight sample space.

The Shapley value for a superpixel $i$, that can be interpreted as an \emph{importance} score, is defined by
\begin{equation}\label{eq:shapleyValue}
    \phi_i = \sum_{x \in X^\llbr{i}} \Gamma(k-1, |x|) \bigl(f(\xi_{x[i\gets 1]}) - f(\xi_x) \bigr)
\end{equation}
with $X^\llbr{i}$ being the set of all masks $x$ having $x[i]=0$, and with the \emph{Shapley importance} function\,\cite[p.\,6]{monderer2002variationsShapley}
\begin{equation}
    \Gamma(k, |x|) = \frac{1}{(k+1) \binom{k}{|x|}}
\end{equation}
Fig.\,\ref{fig:binomShapley}B shows the Shapley importance function for a few $k$ values.
Higher values of $\Gamma(k, |x|)$ for a mask $x$ means that samples having that mask will weight more in the final value of $\phi_i$. 
Comparing Fig.\,\ref{fig:binomShapley}A and B clearly shows that LIME Image samples the majority of the masks among those having the least importance (in the Shapley sense).
In fact when $p=0.5$ it holds that
\begin{align*}
    \mathcal{B}(k, |x|) \cdot \Gamma(k, |x|) &=
    \frac{\binom{k}{|x|} p^{|x|} (1-p)^{k-|x|} }{(k+1) \binom{k}{|x|}}
    =\frac{0.5^k}{k+1}
\end{align*}
i.e. the Shapley importance is the reciprocal (times a constant) of the binomial distribution $B(0.5)$ used by LIME.
This is an informative detail of the Shapley theory, which motivates the proposed sampling theory.

Interestingly, Shapley value computation is not typically performed as a Monte Carlo sampling, but adopts other strategies to generate the samples\,\cite{okhrati2021multilinear,mitchell2022sampling}. 
For instance, in \cite{owen1972multilinear} Eq.\,\eqref{eq:shapleyValue} is rewritten as
\begin{equation}\label{eq:owen}
    \phi_i = \int_0^1 \Biggl( 
        \sum_{x \in X^\llbr{i}_q} \frac{1}{|X^\llbr{i}_q|} 
            \bigl(f(\xi_{x[i\gets 1]}) - f(\xi_x) \bigr) 
        \Biggr) \,\mathrm{d}q
\end{equation}
with $X^\llbr{i}_q$ being a random subset of masks $x$, having $x[i]=0$ and, for all $j \neq i$, $x[j] \sim B(q)$ with $B(q)$ a Bernoulli-distributed random variable having probability $q$.
Such strategy allows to get samples across the entire spectrum of $|x|$ values. 
In the rest of the paper we shall discuss a strategy for LIME Image where $x$ values are not sampled from $B(0.5)$ as in Eq.\,\eqref{eq:maskB} but from a modified version of Eq.\,\eqref{eq:owen}.


\section{Proposed Methodology}

We describe a methodology based on stratified sampling of the $X$ values, where each stratum has a uniform probability of being selected and represented in the samples of $X$.
This oversamples the ``rare'' samples at the tail of the $Y$ distribution, improving the samples over which the linear regressor is fit.
However, this sampling could result in a form of \emph{bias}.
To avoid that, an adjustment factor is introduced to counterbalance the oversampled data points.

Let $\popX$ denote the complete population of mask samples, having $2^k$ elements,
 and let $\popY$ be the dependent variable of $\popX$.
Consider a stratified partitioning.
Let $\popX^{(i)}$ be the set of all possible masks having $|x|=i$, i.e. for which exactly $i$ superpixel are preserved.. 
Clearly, $\popX^{(0)} \cdots \popX^{(k)}$ forms a partitioning of all possible masks, and
\begin{equation*}
    \{0,1\}^k = \bigcup_{i=0}^k \popX^{(i)}
\end{equation*}
since any possible mask $x$ appears in one (and only one) set $\popX^{(|x|)}$. 
Moreover $\popX^{(0)} {=} \{\vec{\mathbf{0}}\}$ and $\popX^{(k)} {=} \{\vec{\mathbf{1}}\}$ 
(masks for the explained input sample with everything/nothing perturbed, resp.).
Each stratum $\popX^{(i)}$ does not have a uniform number of samples, but its size is known a-priori since they follow the binomial distribution, i.e.
\begin{equation}\label{eq:stratumSize}
    |\popX^{(i)}| = \binom{k}{i},    \qquad 0 \leq i \leq k
\end{equation}

In an unbiased Monte Carlo sampling model, as Eq.\,\eqref{eq:maskB}, the probability of selecting a sample $x$ in a from stratum $\popX^{(i)}$, with $i=|x|$, is therefore proportional to that stratum probability in the overall population $\popX$, i.e.
\begin{equation*}
    \mathit{Prob}\bigl\{x \in \popX^{(i)} ~|~ x \in X \bigr\} = 
            \frac{|\popX^{(i)}|}{\sum_{j=0}^k |\popX^{(j)}| } =
            \frac{\binom{k}{i}}{2^k}
\end{equation*}

Let $\overX$ be an oversampled population, where the probability of taking samples from any of the $k+1$ strata is uniform, and does not depend on the stratum size, i.e.
\begin{equation*}
    \mathit{Prob}\bigl\{x \in \popX^{(i)} ~|~ x \in \overX \} = 
        \frac{1}{k+1}
\end{equation*}
Let $\overY$ be the corresponding dependent variables for $\overX$. 
We can derive an \emph{adjustment factor} for the $\overX$ samples to correct the bias introduced by the oversampling, which results for an arbitrary sample $x$ in stratum $\popX^{(i)}$ as 
\begin{equation} \label{eq:adjustment}
    \adjFact(i) = \frac
        {\mathit{Prob}\bigl\{ x \in \popX^{(i)} ~|~ x \in X \bigr\}}
        {\mathit{Prob}\bigl\{ x \in \popX^{(i)} ~|~ x \in \overX \bigr\}} =
        \frac{(k+1)\binom{k}{i}}{2^k}
\end{equation}
Weighted regression with the oversampled set $\overX$ can be obtained by inserting the adjustment factor as a multiplicative term in the existing weight equation of LIME.
Let $\ovw_{\ovx}$ be the weight of sample $\ovx \in \overX$ obtained from Eq.\,\eqref{eq:weight} multiplied by $\adjFact(|\ovx|)$, and let 
$\overW = \{ \ovw_{\ovx} ~|~ \ovx \in \overX \}$ be the set of weights for the set $\overX$.
Then let
\begin{equation} \label{eq:linRegStrat}
    \overBeta = ({\overX}^\textsf{T} \overW \overX)^{-1} {\overX}^\textsf{T} \overW \overY
\end{equation}
be the weighted least square estimator of the regression coefficients of $\overY$ on $\overX$ that takes into account the strata density of the oversampled set $\overX$.

\paragraph{The Mixture Model.}
The linear homoscedastic regression model of Eq.\,\eqref{eq:linReg} adopted by LIME may not be particularly accurate when strata at the tails are severely undersampled, and these strata are significantly different from the mean.
In that case, $\beta$ is not globally unique across the sampled population, but varies by stratum
\begin{equation}\label{eq:mixtureModel}
    {\overY}^{(i)} = {\overX}^{(i)} \cdot {\overBeta}^{(i)} + {\widehat{\epsilon}}^{(i)}
\end{equation}
Intuitively, the ${\overBeta}^{(i)}$ vectors represents the feature importance for stratum $i$, which is at uniform ``distance'' from the input sample $\xi$. 
The closer $i$ is to $k$, the closer $\xi_x$ is to $\xi$.

\paragraph{Impact of Stratified Sampling in LIME Image.}
The impact of using a weighted regression from a stratified sampling schema may not be negligible. We simplify the analysis considering two cases.

\medskip
\noindent\textbf{Case (A):}
The mean and variance of $\overBeta^{(i)}$ are independent of the strata (i.e. the population structure is \emph{homoscedastic}).
Then it is easy to see that $\mathbb{E}[\beta] \approx \mathbb{E}[{\overBeta}^{(i)}]$, for any $i$.
In that case, a weighted regression model of Eq.\,\eqref{eq:linRegStrat} is not needed, and the model computed by LIME using Monte Carlo sampling will not have issues due to the undersampling of the tails.
In that case, the stratified sampling will converge to the same values, regardless of the strata ratios in the synthetic neighborhood.

\medskip
\noindent\textbf{Case (B):}
The mean and variance of $\overBeta^{(i)}$ varies by stratum.
In that case, the bias introduced by the Monte Carlo sampling scheme will not allow to consider the systematic differences in the stratum, and a weighted regression or a mixed model built on a stratified sampling strategy are highly advisable\,\cite{dumouchel1983using}.
\\
In a  certain sense Case (B) is even worse, because the undersampling of the neighborhood of $\xi$ breaks the logic of building models that are locally faithful to the black box model $f$ in the neighborhood of the explained sample, since the local neighborhood (close to $\xi$) that is really representing the local behaviour is missing/undersampled.


\begin{algorithm}[t]
\caption{Neighborhood sampling strategies}
\label{algo:sampling}
\Function{MonteCarloSampling$(n,k)$}{
\nl  $X \gets n \times k$ matrix \;
\nl  \For{$i$ between 1 and $n$} {
\nl    \For{$j$ between 1 and $k$} {
\nl      $X[i,j] \gets B(0.5)$
       }
     }
   }
\setcounter{AlgoLine}{0}
\Function{StratifiedSampling$(n,k)$}{
\nl  $X \gets n \times k$ matrix \;
\nl  \For{$i$ between 1 and $n$} {
\nl    $q \gets \mathit{Uniform}(0, 1)$ \;
\nl    \For{$j$ between 1 and $k$} {
\nl      $X[i,j] \gets B(q)$ \;
\nl      $\adjFact[i] \gets (k+1) \cdot \frac{1}{2^k} \cdot \binom{k}{|X[i]|}$ 
       }
     }
   }
\end{algorithm}

\medskip
Algorithm \ref{algo:sampling} outlines two sampling methods: the original Monte Carlo sampling used by LIME Image, and the introduced stratified sampling technique.
The \emph{MonteCarloSampling} function computes the data matrix $X$ (from Eq. \ref{eq:maskB}) with replacement.
Function \emph{StratifiedSampling} is one possible way of generating a stratified population, similarly to Eq.\,\eqref{eq:owen}.
For every sample $i$, a single coefficient $q$ is 
randomly drawn from a uniform distribution ranging between $0$ and $1$. 
The individual values of the $i$-th mask vector are then sampled from a Bernoulli random variable $B(q)$ with probability $q$. 
This will obtain a sample $X[i]$ in stratum $\overX^{(|X[i]|)}$, where strata have now equal probability of being selected.
The adjustment factor $\adjFact[i]$ for sample $i$ is also computed.

Other strategies could also be employed\,\cite{neymanAllocation}.
An interesting approach suggested in \cite{konijn1962regression} for computing the coefficients would be to fit one linear regressor for every strata and then form a \emph{mixed model} with the coefficients' averages. 
This approach however requires more changes in the LIME code, thus we have favored the approach of Algorithm\,\ref{algo:sampling} which is more straightforward.

\section{Experimental Evaluation} \label{sec:experimental}

We perform experiments to compare the proposed methodology with the original Monte Carlo setup of LIME, in order to test whether the generated distributions of $\overY$ have a better sampling, resulting in feature attribution vectors $\overBeta$ that are less confused.

\begin{figure}[t]
    \centering
    \includegraphics[width=1.0\columnwidth]{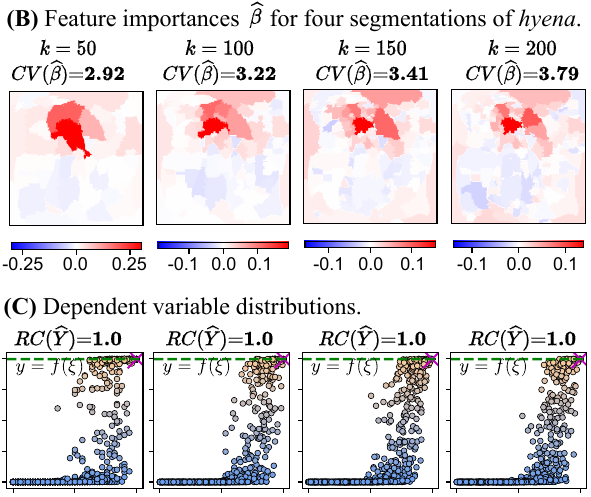}
    \caption{Four explanations $\overBeta$ of the same image of Fig.\,\ref{fig:hyenaMC} using stratified sampling (each is an average of $10$ runs).}
    \label{fig:hyenaStrat}
\end{figure}

We start by revisiting the \emph{hyena} example of Fig.\,\ref{fig:hyenaMC} but recomputed using the \emph{StratifiedSampling} algorithm. The results are reported in Fig.\,\ref{fig:hyenaStrat}.
The first thing to observe is that the dependent variable distribution has now samples for several different classification scores, which allows the linear regressor to be fit against a synthetic neighborhood with better variation than in the standard Monte Carlo setup of Fig.\,\ref{fig:hyenaMC}B.
The heatmap of the explanations also reflect this improvement: feature attribution values now have a much better coefficients of variation, resulting is some superpixels receiving high importance, and other receiving almost zero importance. Moreover, the explanation remains reasonably consistent, identifying the same ``spot'' in the image even when the set of superpixels changes.
Moreover, Fig.\,\ref{fig:hyenaStrat}C shows that the distribution of the dependent variable (the $y$-axis) across the strata (the $|x|$ value on the $x$-axis) is far from being homoscedastic. This further reinforces the need for stratified sampling in the process.

\begin{figure}[t]
    \centering
    \includegraphics[width=\columnwidth]{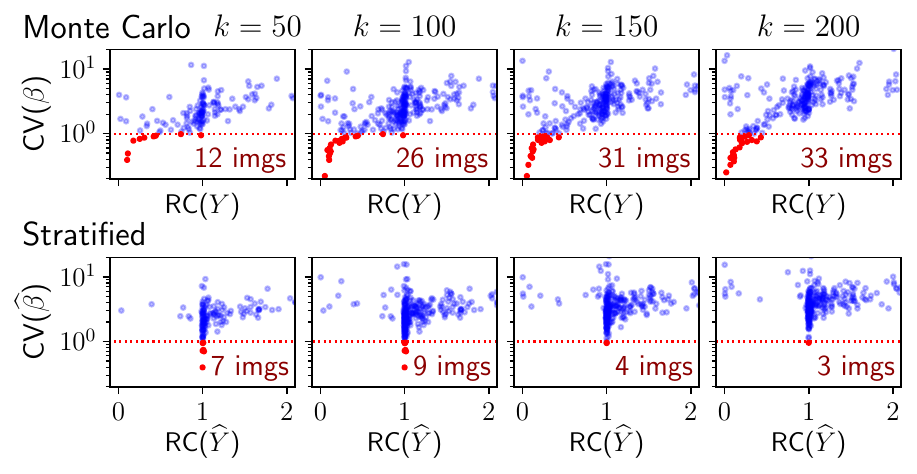}
    \hrule
    ~~~~~~~~\includegraphics[width=0.92\columnwidth]{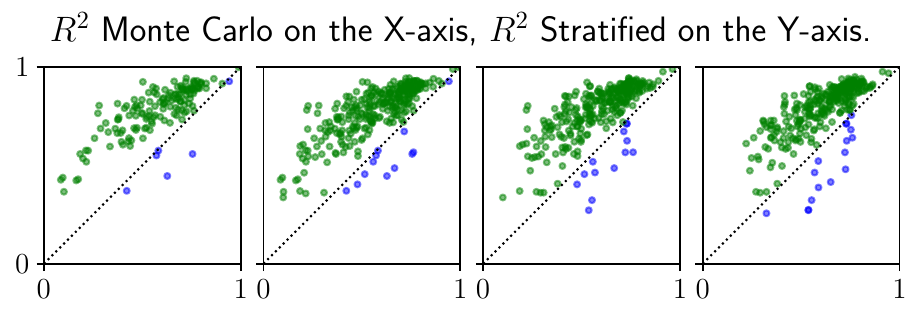}
    \caption{$CV$ vs $RC$ and $R^2$ comparisons, for 150 images.}
    \label{fig:MCvsStrat:YvsCV}
\end{figure}

To better quantify the effect, we took the first 150 images of the \textit{ImageNet Object Localization dataset}\,\cite{imagenet-object-localization-challenge}.
For each image we performed a dichotomic search on the \emph{max\_dist} hyper-parameter to find a configuration of \emph{quick shift} that results in a number of superpixels $k$ equal to $50$, $100$, $150$ and $200$.
For each range, we run 10 times LIME Image with both the Monte Carlo and the stratified sampling using $n{=}1000$ samples, and record both the average range coverage $RC$ of the $Y$ ($\overY$ resp.) distributions and the $CV$ of the feature attribution vectors $\beta$ ($\overBeta$ resp.).
The first two rows of plots in Fig.\,\ref{fig:MCvsStrat:YvsCV} show the results obtained from Monte Carlo (above) and stratified sampling (below). Each plot has 150 dots, one for each image in the dataset for a fixed $k$.
Each dot has the $CV$ on the $y$-axis, and the range coverage $RC$ on the $x$-axis. 
It is very clear that the stratified sampling approach ensures that the range of $\overY$ distribution range is well covered w.r.t. the $Y$ distribution.
At the same  time, the Monte Carlo approach produces, for some images, explanations with very poor variation in the coefficients, and this is clearly linked with the low range coverage.
Explanations with an average $CV$ below one are highlighted.
The third row in Fig.\,\ref{fig:MCvsStrat:YvsCV} reports the comparison of the average $R^2$ coefficients for the Stratified (on the $y$ axis) and for the Monte Carlo (on the $x$ axis), showing that, on average, the $\overY$  distribution better explains the $X$ distribution than $Y$.

\begin{figure}[t]
    \centering
    \includegraphics[width=\columnwidth]{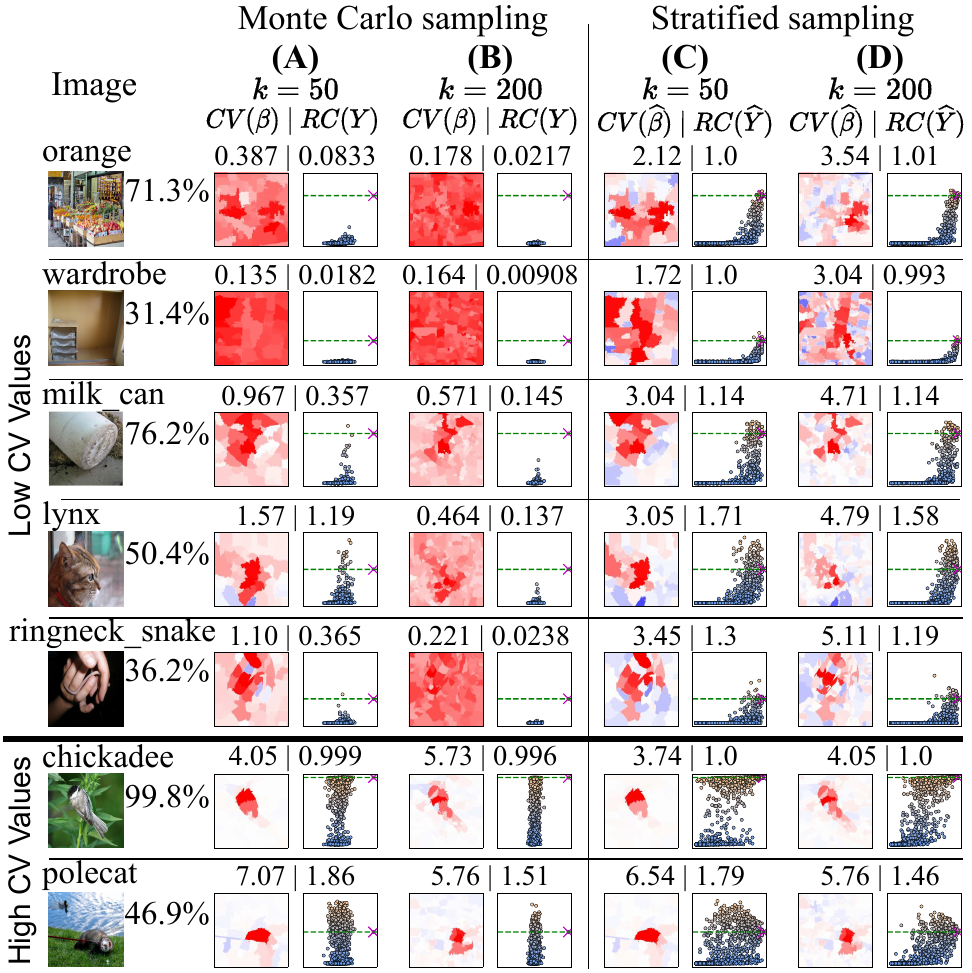}
    \caption{Examples of LIME Image explanations in the lower-left tail of Fig.\,\ref{fig:MCvsStrat:YvsCV}, with heatmaps, $CV$ and $RC$ values.}
    \label{fig:severalExamples}
\end{figure}

We report some of these images with low $CV$ values in Fig.\,\ref{fig:severalExamples} (first five rows). 
Columns A and B show the Monte Carlo sampling, C and D the Stratified sampling. 
We consider the cases with $k{=}50$ (columns A and C) and $k{=}200$ (columns B and D). 
For each explanation we show the heatmap and the $Y$ ($\overY$ resp.) distribution, together with the $CV$ and $RC$ values.
Column B clearly shows the problem: the Monte Carlo sampled distributions are very poor, with all $Y$ almost close to $0$.
This results in feature attribution vectors $\beta$ that are almost uniform, which do not identify any relevant sub-region of the explained images. 
This detrimental effect is greatly reduced by the stratified sampling approach, which remains capable of identifying a sub-region of the image that is deemed to be responsible for the classification.
When the sampled distribution is sufficient, both the Monte Carlo and the Stratified sampling approaches converge to similar explanations (last 2 rows of Fig.\,\ref{fig:severalExamples}).

\section{Conclusions}

We have provided a reformulation of the sampling strategy of LIME Images  showing its critical role in cases where the simple linear homoscedastic model for regression is not true, i.e. when the $Y$ value are undersampled by a Monte Carlo strategy. 
This happens when the black-box model $f$ (almost always) returns low classification scores when about $\sim 50\%$ of the explained image $\xi$ is masked, resulting in flat $Y$ distributions with very low range coverage, for which the coefficient $\beta$ of a linear regression model will be close-to constant (i.e. with low variation).
We considered image data, using the popular ImageNet dataset for the experiments.
Of course the strategy could be of interest for other kind of data, even if some adjustments are probably needed (since the interpretable feature space for images is over the booleans, unlike for other data types). 
Moreover, a more extensive test could be useful to assess its applicability.

We focused on reformulating the regression strategy of LIME.
Observations from the Shapley theory suggests that another formulation that gives uniform weight to all strata is also possible, but it was not considered in this paper, and further investigations are needed.
The goal of the proposed methodology is to avoid the undersampling of $Y$.
In addition, the work of\,\cite{haberman1975much} proves various results and bounds between $\beta$ and $\overBeta$, which could be explored further.
The formulas were formulated assuming no regularization factor: however, since the main changes are in the sampling strategy, it should be possible to extend these results to ridge regression.
The (briefly introduced) mixed model could also be used instead of randomly selecting the strata from a uniform distribution in the proposed algorithm.
As a future work, we plan to reformulate LIME equations to better follow the neighborhood locality, which is not captured by sampling from the binomial distribution, as described in the "Limitations" section and illustrated in Fig.\,\ref{fig:limeNeighborhood}.

\paragraph{Availability}
The LIME Image with stratified sampling is available at: 
\url{https://github.com/rashidrao-pk/lime_stratified}
All code needed to replicate the experiments (including the \emph{requirements.txt} with the library versions used) can be found at: 
\url{https://github.com/rashidrao-pk/lime-stratified-examples}


\clearpage

\section{Acknowledgments}

This work has received funding from the European Union's Horizon 2020 research and innovation program  ECSEL Joint Undertaking (JU)  under Grant Agreement No. 876487, NextPerception project. 
The JU receives support from the European Union's Horizon 2020 research and innovation programme and the nations involved in the mentioned projects. The work reflects only the authors' views; the European Commission is not responsible for any use that may be made of the information it contains. 


\bibliography{aaai24.bib}

\end{document}